\documentclass[runningheads]{llncs}
\usepackage{cite}
\usepackage{amsmath,amssymb,amsfonts}
\usepackage{graphicx}
\usepackage{textcomp}
\usepackage{xcolor}
\usepackage{algorithm}
\usepackage{algpseudocode}
\usepackage{caption}

\newcommand{\pluseq}{\mathrel{+}=}

\def\BibTeX{{\rm B\kern-.05em{\sc i\kern-.025em b}\kern-.08em
    T\kern-.1667em\lower.7ex\hbox{E}\kern-.125emX}}
\begin{document}

\title{Packet2Vec: Utilizing Word2Vec for Feature Extraction in Packet Data
}


\author{Eric L. Goodman \and
Chase Zimmerman \and
Corey Hudson}
\authorrunning{E. Goodman et al.}

\institute{Sandia National Laboratories, Albuquerque, NM, USA}

\maketitle

\begin{abstract}
One of deep learning's attractive benefits is the ability to
automatically extract relevant features for a target problem
from largely raw data, instead of utilizing human engineered
and error prone hand-crafted features.  While deep learning
has shown success in fields such as image classification
and natural language processing, its application
for feature extraction on raw network packet data for 
intrusion detection is largely unexplored.  In this paper
we modify a Word2Vec approach, used for text processing,
and apply it to packet data for automatic feature extraction.
We call this approach \emph{Packet2Vec}.
For the classification task of benign versus malicious traffic
on a 2009 DARPA network data set,  we obtain an area under the 
curve (AUC) of the receiver operating characteristic (ROC) 
between 0.988-0.996
and an AUC of the Precision/Recall curve between 0.604-0.667.
\end{abstract}

\section{Introduction}
\label{section:introduction}
An appealing aspect of many deep learning approaches is the
ability to automatically extract features 
from largely unprocessed
data.  In Krizhevsky et al. \cite{NIPS2012_4824}, one of the seminal 
works that started the popularization of convolutional neural networks
applied to images, they show that the learned 
early convolutional kernels displayed a range of image 
filters, similar to hand-crafted features from
more traditional vision processing approaches 
such as SIFT \cite{SIFT1999} and SURF \cite{SURF2008}.  

For text processing, Word2Vec approaches 
\cite{word2vec-2013-1, word2vec-2013-2} create a vectorized 
representation of words, called
embeddings, where similar words (e.g. King and Queen)
are close distance-wise in the embedded space.  
Vector operations also make intuitive sense, such as 
\emph{King - Man + Woman = Queen}, meaning that the vector representation
of \emph{King} minus the vector for \emph{Man} plus the vector for 
\emph{Woman} creates a vector where the closest word embedding is the one
for \emph{Queen}.  This feat is achieved on a large corpus of raw text with
little to no-preprocessing.  The deep learning approach is able to 
create these word embeddings just based on the text itself without
human-engineered feature extraction.

Cyber data and intrusion detection is an area ripe for exploration
of how deep learning can automatically extract features from raw
packet data.  However, most of the current work applying deep learning
to intrustion detection relies upon the features already being extracted from
packet data \cite{rnn-ids-2017, Javaid:2016:DLA:2954721.2954780}. 
Many researchers choose to
use data sets such as NSL-KDD \cite{NSL-KDD2009, NSL-KDD2013} 
or the original 1999 KDD data set, both of which have 
41 features to represent the network packet data.

Instead of creating hand-crafted features for 
each packet, the approach we take is to pass the 
raw packet data through a Word2Vec approach to create
a vectorized representation for each packet, and then perform 
classification of the packet based on that representation.

Specifically, our approach has the following steps:
\begin{itemize}
\item \textbf{N-grams:} Word2Vec requires a sequence of tokens.  
Packet data has no clear analog.  To address this, we take each
packet and transform it into a sequence of n-grams.  
This forms our sequence of words, similar to the presentation of text.
We purposefully throw out IP and port information, as we want
the representation of the packet to be based on content, not
who sent it.

\item \textbf{Embeddings:} Once we have a sequence
of n-grams, applying Word2Vec is straightforward, and we create
a vectorized representation for each frequent n-gram (vocab size is a 
hyperparameter).

\item \textbf{Feature Vectors:} To perform classification
on each packet, we need a fixed-size vector representation for each
packet.  We take the simple approach of averaging the
word embeddings for all of the n-grams in a packet, i.e.
\begin{equation}
v(p) = \frac{\sum_{t \in p} e(t)}{|p|}
\end{equation}
where $p$ is a packet, $t \in p$ are the n-grams of $p$, $|p|$ is 
the number of n-grams found in $p$, and $e(t)$ returns the embedding
for n-gram $t$.

\item \textbf{Learning and Classification:} Once we have each packet 
translated into fixed-size feature vectors, we then pass those feature
vectors to a supervised machine learning approach for training
and then testing on unseen data.
\end{itemize}

Intrusion detection is an important area of research, vital
for protection of national infrastructures, intellectual property, 
financial systems, privacy, and safety; however, the problem is a moving 
target, an arms race between defenders and attackers, along with
constant evolution of the underlying technologies.   
There is evidence of growing 
sophistication among malicious actors.  Symantec reports that the number
of targetted attack groups, i.e. groups that are professional, 
highly organized, and target specifically rather than
indiscriminately, grew at a rate of 29 groups a year between the years of
2015 to 2017, from a total of 87 to 140 \cite{istr-2018}.  
Also, as evidence of constant change in the cyber arena, 
the number of IoT (Internet of Things) attacks grew 
by 600\%, an increase of 54\% of mobile
malware variants, and an 80\% increase in Mac malware.

We view our contribution as a way to increase the rate that defenders
can evolve their methods to protect networks and infrastructure.  Instead
of manually hand-crafting features, which is error prone and difficult
to determine impact, we can rely upon our Packet2Vec approach to
automatically calculate features of interest.  

The rest of this paper is organized as follows: 
Section \ref{section:approach} describes our approach in
detail, including the steps we took to parallelize our solution.
Section \ref{section:results} presents the results of
using our approach on a large cyber data set.
Section \ref{section:related} covers related work.
Section \ref{section:conclusions} concludes.

\section{Approach}
\label{section:approach}

In the introduction, we presented our 
approach at a high-level.  However,
applying Word2Vec on cyber data is challenging 
due to amount of information.
In particular, we examined the DARPA 2009 data set \cite{DARPA2009}.
This data set spans a period of 10 days, 
from November 3rd to November 12th,
2009.  It is broken up into files that are 
just over 1 billion bytes (954 MBs),
where each file represents 1-6 minutes worth of 
traffic. In this work we
examined the first day, which is roughly 15.5 hours 
(it starts after 8:30 am) and 
comprises 558.8 GBs in total packet data.  
Due to the size of the data,
we needed to create an iterative process for training our model.

Our solution is a combination of C++ code that is then exposed 
to python using Boost python \cite{boost-python}.  We developed
most of our implementation in C++ for performance, but then exposed
it to python so that we could integrate with the Tensorflow library
\cite{tensorflow2015-whitepaper} for creating the embeddings for 
the n-grams, and the Scikit-learn library \cite{sklearn-api} for
the classifier models to make predictions on whether the packets
are benign or malicious.  We also took efforts to parallelize
the code using standard C++ features such as std::thread to manually
instrument the code. 
As we discuss the implementation, we will
highlight the parallelization.  Also, in Section \ref{section:processing-time},
we will discuss the parallel performance of the code.

\begin{figure*}
\centering
\includegraphics[width=.75\linewidth]{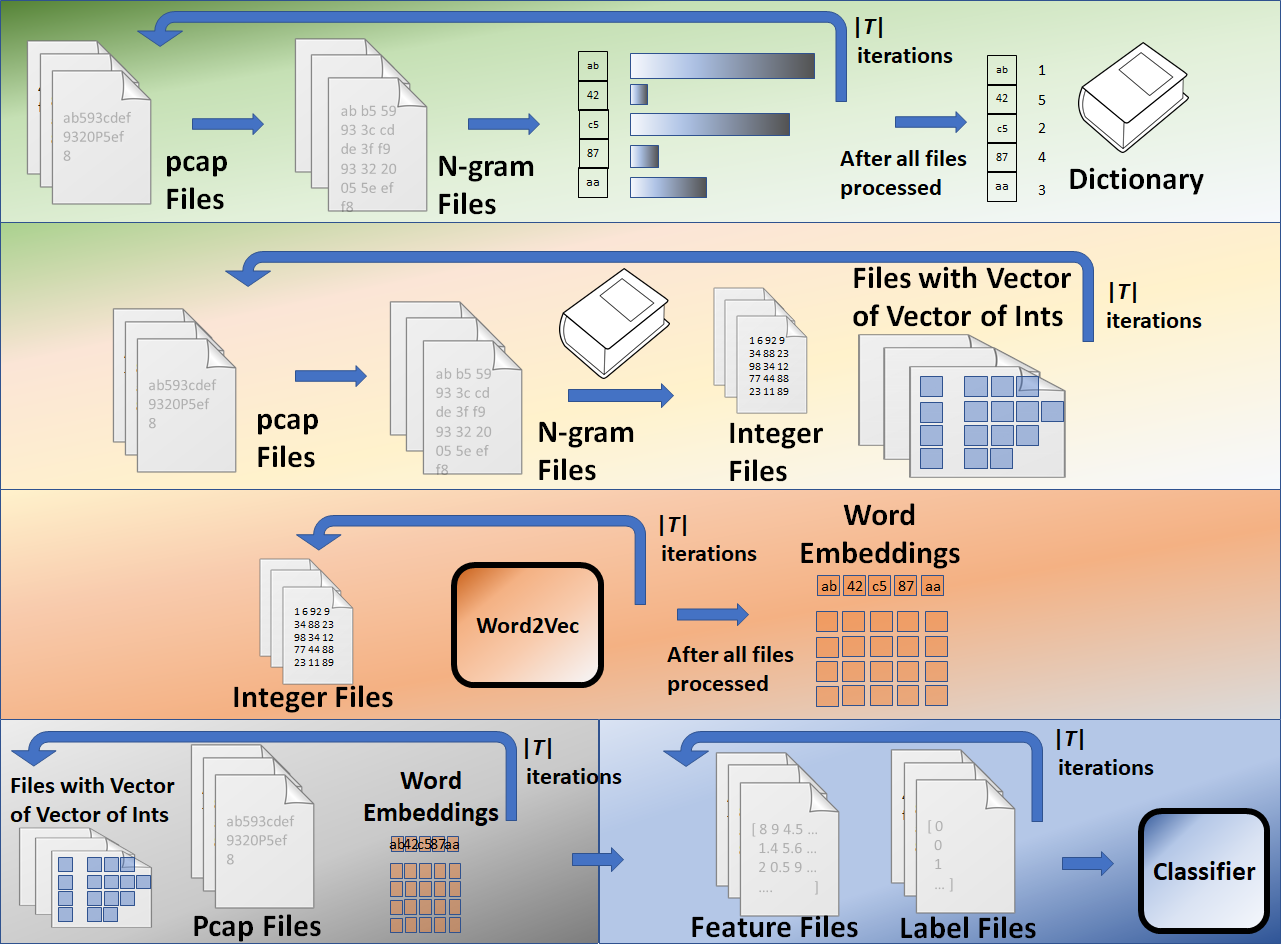}
\caption{Implementation of iterative pcap processing approach. 
The first phase creates a dictionary, mapping n-grams to 
integer identifiers.  The dictionary is utilized in the 
second phase to transform the raw pcap data into integer 
vectors which are saved on disk. In the third phase, a Word2Vec approach 
is applied to the 1D integer vectors to create the n-gram embeddings. 
These embeddings are used in conjunction with the 2D integer vectors 
to create feature vectors (fourth phase) which are then used for training
in the final phase.}
\label{figure:implementation}
\end{figure*}

Figure \ref{figure:implementation} gives an overview 
of the iterative approach. The first phase (pseudocode 
found in Algorithm \ref{algorithm:dictionary}),
creates a dictionary, mapping n-grams to integer identifiers.
The first phase begins by iterating through all pcap 
files used for training, n-gramming each packet, and 
incrementing the counter for each n-gram.  After obtaining 
counts for each n-gram found in all the training files,
identifiers are assigned for the top $|V|$ n-grams, 
where $|V|$ is the size of the vocabulary, a hyperparameter.  
Concerning memory utilization,
we only load one pcap file at a time.  Also, the dictionary is limited
by the number of found n-grams.  We used 2 byte n-grams, which at most
has $2^{16}$ possible values.

\begin{algorithm}
\caption{Training Phase 1: Creating the Dictionary}
\label{algorithm:dictionary}
\begin{algorithmic}[1]
\State $T$, the set of pcap files used for training. 
\State $D$, a dictionary mapping from n-grams to integers.
\State $|V|$, the size of the vocabulary.
\ForAll{$f \in T$}
  \ForAll{$packet \in f$}
    \State $ngrams \gets ngram(packet)$ 
    \ForAll{$ngram \in ngrams$}
      \State $D[ngram] \gets D[ngram] + 1$
    \EndFor
  \EndFor
\EndFor
\State $keys \gets sort(D)$ \Comment{Keys sorted by decreasing frequency} 
\State $D.clear()$ \Comment{Clear out counts}
\State $i \gets 1$
\While {$i \le |V|$}          \label{algorithm:dictionary:line:while}
  \State $D[key[i]] \gets i$
  \State $i \gets i + 1$
\EndWhile                     \label{algorithm:dictionary:line:endwhile}
\State $write(D)$             \label{algorithm:dictionary:line:write}
\end{algorithmic}
\end{algorithm}

The actual implementation of Algorithm \ref{algorithm:dictionary}
is a bit more nuanced as we structured it in such a way to enable
parallelization.  We first iterate in parallel over all packets and
n-gram them.  This is embarrassingly parallel and 
requires no inter-thread coordination.  The end result 
is a vector of vector of n-grams.  Then we flatten the vector 
of vector of n-grams into a single vector of n-grams,
again in parallel.  Finally, we hand the single vector of all n-grams
to the dictionary, which updates the frequency counts for each n-gram.
This is the only loop that requires coordination between threads,
as two threads can potentially try to update the count for the same
n-gram; however, adding mutexes around the update routine makes it 
thread safe.  After all files have been processed, 
we also parallelize the implementation
of lines \ref{algorithm:dictionary:line:while} - 
\ref{algorithm:dictionary:line:endwhile}. 
We need the dictionary for later phases, so we write it out to disk
on line \ref{algorithm:dictionary:line:write}.

The second phase (Algorithm \ref{algorithm:translate}) 
utilizes the dictionary created from the first phase
to translate the pcap files into integers.  We iterate through
each pcap file (line \ref{algorithm:translate:for1}), 
creating two data structures for each pcap file.  
One data structure is a list of integers
(line \ref{algorithm:translate:list}), which is the pcap file
translated into integers using the dictionary.  There is also
a vector of vector of integers (line \ref{algorithm:translate:vv}),
which is the same as the integer list, but
now indexed by packet.  After processing a pcap file, we write out
the list of integers (line \ref{algorithm:translate:write1}) 
and the vector of vector of integers to disk 
(line \ref{algorithm:translate:write2}).
This again allows us to process all of the large pcap files without
exceeding memory limits.  We also parallelize the for loop of
line \ref{algorithm:translate:for1}.  Each of the packets
can be handled independently, so it is embarrasingly parallel.

The third phase is where we create the word embeddings, i.e. 
vectorized representations for each n-gram in the vocabulary. 
The process is described in Algorithm \ref{algorithm:embeddings}
in high level pseudocode.  We iterate over all the integer
files (pcap files translated by the dictionary into a single 
sequence of integers).  On the first iteration we create an 
embedding model based on the first integer file using a
standard word2vec approach.
This creates a matrix of size $|V| \times embedding\_size$,
where each row corresponds to the learned vector representation of
an n-gram.  This first embedding matrix serves as the starting
point for the next iteration of applying word2vec to another
integer file.  We continue in this manner until all integer 
files have been processed.

\noindent \begin{minipage}{\textwidth}
\centering
\begin{minipage}{.45\textwidth}
\begin{algorithm}[H]
\caption{Training Phase 2: Translating Pcap Files}
\label{algorithm:translate}
\begin{algorithmic}[1]
\ForAll{$f \in T$}                         \label{algorithm:translate:for1}
  \State $l$, a list of integers \label{algorithm:translate:list}
  \State $vv$, a 2D vector of integers \label{algorithm:translate:vv}
  \ForAll{$packet \in f$}  \label{algorithm:translate:for2}
    \State $v$, a vector of integers
    \State $ngrams \gets ngram(packet)$
    \ForAll{$ng \in ngrams$}
      \State $l.push\_back(D[ng])$
      \State $v.push\_back(D[ng])$
    \EndFor
    \State $vv.push\_back(v)$
  \EndFor
  \State $write(l)$  \label{algorithm:translate:write1}
  \State $write(vv)$ \label{algorithm:translate:write2}
\EndFor
\end{algorithmic}
\end{algorithm}
\end{minipage}
\hspace{.05\textwidth}
\begin{minipage}{.45\textwidth}
\begin{algorithm}[H]
\caption{Training Phase 3: Creating Word Embeddings}
\label{algorithm:embeddings}
\begin{algorithmic}[1]
\State $L$, the set of files with lists of integers
\State $first\_run \gets True$
\ForAll{$l \in L$}
  \If {$first\_run$}
    \State{$E \gets create\_model(l)$}
    \State $first\_run = False$
  \Else
    \State{$E \gets update\_model(l, E)$}
  \EndIf
\EndFor
\State $write(E)$ \label{algorithm:embeddings:write}
\end{algorithmic}
\end{algorithm}
\end{minipage}
\end{minipage}

The method for training the model is a standard word2vec
approach.  We use the skip-gram model \cite{word2vec-2013-2} 
with noise constrastive estimation \cite{NCE-2012}.
The basis of this approach is for the network to predict
the context given a target word.  However, with noise
constrastive estimation, it becomes a logistic regression
problem where the network is making a binary classification
for each word in the vocabulary of whether or not it came the distribution
of context words or from the noise distribution (unrelated words).
The hyperparameters associated with this approach include the following.
In paranthesis we specify the value we used in our experiments.
\emph{Batch size} (128): The number of words considered at one time.
\emph{Skip window} (1): How big of a context window to consider.  
A value of one selects words to the left and right of the target word.
\emph{Num skips} (2): The \emph{batch size} is divided by 
\emph{num skips} to determine the number of skip windows. 
\emph{Embedding size} (128): The size of each embedding vector.
\emph{Num negative} (64): The number of negative examples used per batch.
\emph{Num steps} (100000): How many batches to create and from which to train.

The fourth phase utilizes the word embeddings in conjunction with
the two dimensional integer vectors to create the feature files.
Each feature file is a matrix where each row represents the features
derived for a packet.  On line \ref{fv:for1} we iterate over
the two dimensional integer vector files, $VV$.  On line \ref{fv:for2}
we iterate over each vector, $v$, within the two dimensional integer
vector, $vv$.  $v$ is a vector of integers, representing the n-grams
of the original packet translated using $D$, the dictionary from 
Algorithm \ref{algorithm:dictionary}.  To create a single
representation for the entire packet, we use the simple 
strategy of averaging the embeddings
(lines \ref{fv:average1} - \ref{fv:average2}).  In the end,
we write out each feature matrix, $X$, to disk (line \ref{fv:writex}).

There is also another process for producing labels for the data.  
The DARPA-2009 dataset has a spreadsheet with labels; however,
the labeling is not at the individual packet level.  
It lists times, IP addresses,
and ports used by malicious traffic.  Thus, to create labels, we
read in the original pcap files and evaluate each packet, checking
if the parameters of the packet match those of an entry in the label
spreadsheet.

\noindent \begin{minipage}{\textwidth}
\centering
\begin{minipage}{.45\textwidth}
\begin{algorithm}[H]
\caption{Training Phase 4: Create Feature Vectors}
\begin{algorithmic}[1]
\State $VV$, set of files with 2D vector of integers 
\State $E$, the word embeddings indexed by integer identifier
\For {$i \gets 1$ to $|VV|$} \label{fv:for1}
  \State $vv \gets VV[i]$
  \State $X$, a matrix of features
  \For {$j \gets 1$ to $|vv|$} \label{fv:for2}
    \State $v \gets vv[j]$
    \State $x$, a vector of features 
    \ForAll{$integer \in v$} \label{fv:for3} \label{fv:average1}
      \State $x \gets x + E[integer]$
    \EndFor
    \State $x \gets x / |v|$ \label{fv:average2}
    \State $X[j] \gets x$ 
  \EndFor
  \State $write(X)$ \label{fv:writex}
\EndFor
\end{algorithmic}
\end{algorithm}
\end{minipage}
\hspace{.05\textwidth}
\begin{minipage}{.45\textwidth}
\begin{algorithm}[H]
\caption{Training Phase 5: Train Classifier}
\label{algorithm:classifier}
\begin{algorithmic}[1]
\State $X_{files}$, the list of feature files.
\State $y_{files}$, the list of label files.
\State $n_{est}$, the number estimators per file.
\State $rfc \gets RFC(warm\_start=True,n_{est})$ \label{train:warm_start}
\State $i \gets 0$
\For {$i \gets 1$ to $|X_{files}|$} 
  \State $X \gets X_{files}[i]$
  \If {$X$ has positive} \label{train:contains}
    \If {$i \neq 0$}
      \State $rfc.n_{est} \pluseq n_{est}$
    \EndIf
    \State $y \gets y_{files}[i]$
    \State $rfc.fit(X, y)$
  \EndIf
\EndFor
\State $write(rfc)$ \label{train:write}
\end{algorithmic}
\end{algorithm}
\end{minipage}
\end{minipage}

The last phase of training is to train an actual classifier.  After phase 4,
we finally have the data in a format that can be ingested by a standard machine
learning algorithm.  We have a set of files that contain the feature vectors
for each packet, and we have another corresponding set of files that have
a binary label indicating a benign/malicious packet.   
Algorithm \ref{algorithm:classifier} outlines the iterative approach to
learning.  In particular we show pseudocode related to the Random
Forest Classifier \cite{random-forest-2001}, 
but it can be easily generalized to other
machine learning algorithms.  An important point to note here
is the \emph{warm\_start} parameter on line \ref{train:warm_start}.
Since we are training in batches over many files, we need to maintain
what was learned from earlier files.  The \emph{warm\_start} parameter
of Scikit-learn \cite{sklearn-api} is used when multiple calls to
the \emph{fit} function are used.  In the case of the Random Forest
Classifier, a number of estimators (trees) are created per file.
However, this doesn't work if a file does not contain any malicious
examples.  On line \ref{train:contains} we skip any files
that do not have malicious packets.
What \emph{warm\_start} means differs depending on 
the classifier used.  For example, with neural networks 
we would initialize the model with the weights learned 
from training on previous files.

\section{Results}
\label{section:results}

In this section we discuss two aspects of performance: 
1) the throughput achieved when applying a trained classifier,
and 2) the classifier performance in detecting malicious network 
activity.
The system we used for our experiments was a DGX \cite{DGX}, 
a supercomputer designed for accelerating deep learning
applications with powerful GPUs.  However, except for the
Packet2Vec portion that creates embeddings, 
our code primarily uses the CPU.  The CPU is a 
dual Intel Xeon 20-core E5-2698 v4 2.2 GHz processor with
512 GBs 2133 MHz DDR4 memory.  There is some variability
to the timing of runs as other users are also using the
system concurrently.

We tested our implmentation on the DARPA-2009 data set 
\cite{DARPA2009}.  DARPA-2009 is a generated data set
covering a period of time from November 3-10, 2009.
Traffic is simulated between a /16 local subnet
that goes through a cisco router to the Internet.
There are a variety of protocols (e.g. HTTP, SMTP, DNS)
and malicious activities (e.g. DDoS, Phishing, 
port scans, spam bots).  For this work, we treat all 
the malicious categories as single class
so the problem is binary classification: malicious or benign.
We evaluated our approach on the first day's worth of data (about
15.5 hours because the data starts around 8:30 am).  In total
for the first day there are 600 pcap files, 
each 1 billion bytes (954 MBs).
Groundtruth labels are provided in the form a spreadsheet
specifying the IPs, ports, and a bounding time window 
of when an attack occurred.
For the portion we used, malicious activity accounted for
0.46\% of the the total packets.

\subsection{Processing Time}
\label{section:processing-time}

In this section we report on the processing time for 
applying a trained classifier on unseen data.  It is 
important that our approach be able to keep pace with data creation.  While application of a trained machine
learning model is generally not a concern - testing 
is often orders of magnitude faster than training - our 
approach does have significant preprocessing steps.  
To classify unseen data, we need the following as input:
1) a pcap file, 
2) the dictionary from n-grams to integers (created 
during Algorithm \ref{algorithm:dictionary} and written to disk
on line \ref{algorithm:dictionary:line:write}),
3) the n-gram embeddings (created from Algorithm
\ref{algorithm:embeddings} and written to disk on line
\ref{algorithm:embeddings:write}), and
4) the trained classifier (created during Algorithm 
\ref{algorithm:classifier} and written to disk on line
\ref{train:write}).


The overall process of applying a trained classifier to unseen data 
is described below.  We will make note of which portions are serial,
serial but could be parallelized, and already parallelized.
\begin{enumerate} 
\item Read pcap object: We read in a pcap object.  Unless there
is parallel I/O, this is largely a serial operation and cannot
be parallelized.
\item N-gram the packets: For each of the packets in the pcap
object, we n-gram them.  This step has been parallelized.
\item Translate the n-grams into integers:  Using the 
dictionary, we translate each vector of n-grams into 
a vector of integers.  This step has been parallelized.
\item Create the feature matrix: This step takes the translated
packet data of integers and converts them into embedding vectors,
averages the embeddings, and then fills a matrix that has
all the feature vectors.  This step should be parallelizable
but since we use a python object within C++ as the feature matrix,
we run into issues with the Python global interpreter lock only
allowing one thread.  This should be surmountable, but will
require a deeper dive into Boost python \cite{boost-python} and 
the NumPy C-API, which is C-based API for manipulating NumPy
data structures (the feature matrix is a \emph{NumPy.ndarray}).
\item C++ to python overhead: The function to create the feature
matrix is written in C++ but we added a python interface.  
The python function reports on average 13.6 seconds more
than the corresponding C++ implementation. We hypothesize this may
be due to memory transfer costs.  Regardless, this will
be difficult to optimize without a deep exploration
into Boost python.
\item Making predictions on the feature matrix:  Here we apply
the trained classifier to the now prepared feature matrix.
We use the Scikit-learn library \cite{sklearn-api} for the
machine learning models.  This step could also be parallelized
using one of the python libraries for parallel execution, but
we have not taken that step yet.
\end{enumerate}

\begin{figure*}
\begin{minipage}{0.5\textwidth}
\centering
\captionsetup{width=0.95\linewidth}
\includegraphics[width=0.8\linewidth]{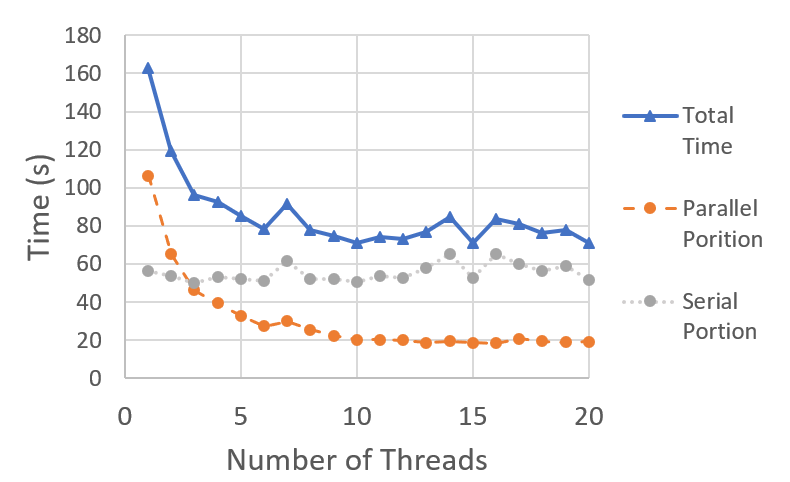}
\caption{Time for Testing One File: We apply a trained Random
Forest Classifier to unseen data and report the times.  The portion
of the code that has been parallelized shows improvement up to
ten threads.}
\label{figure:scaling-times}
\end{minipage}
\begin{minipage}{0.5\textwidth}
\centering
\captionsetup{width=0.95\linewidth}
\includegraphics[width=0.8\linewidth]{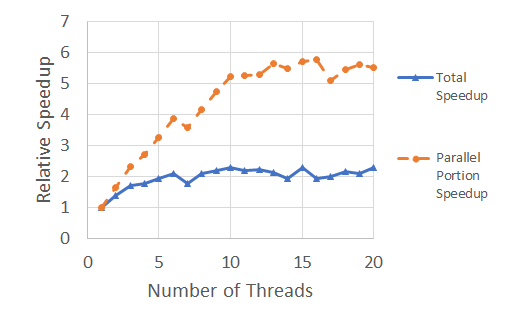}
\caption{Relative Speedup: Same data as Figure 
\ref{figure:scaling-times} but now showing relative speedup
of the overall testing phase and the parallel portion.}
\label{figure:scaling-speedup}
\end{minipage}
\end{figure*}

To evaluate the parallel performance of the pipeline to apply a
trained classifier to unseen data, we trained a 
Random Forest Classifier \cite{random-forest-2001} 
on one pcap file and then tested it on another pcap, varying
the number of threads.  Figure \ref{figure:scaling-times}
gives the overall time while Figure \ref{figure:scaling-speedup}
provides the relative speedup as we increase the thread count.
As expected, the parallel portion's total time decreases 
as we increase the number of threads, though the overall speedup
plateaus around 10 threads.

Since we have good understanding of which portions of the program
are parallel and which are serial, using Amdahl's law we can 
estimate the maximum achievable speedup:
$Speedup(t) = \frac{1}{(1-p) + \frac{p}{t}}$, 
where we can think of $t$ as the number of cores 
applied to the program and $p$ is the proportion
of the code that benefits from parallel execution.
As $t \rightarrow \infty$, the equation becomes just
$Speedup(t) = \frac{1}{1-p}$.
Table \ref{table:speedup} shows the maximum
theoretical speedup based upon the times
from using one thread.  The \emph{Current}
row shows the times for the parallel and serial
portions for our current implementation.  Based
on those numbers, our maximum speedup is about 2.9.
Experimentally, we achieved a 2.3 speedup with
ten threads.  If we parallelized steps 
four and six, which certainly seems possible,
then the maximum speedup is close to 9.2.
Of course this is only single node speedup, and
we can obtain greater aggregate throughput
on a distributed system.
If pcap data is ingested on multiple nodes,
the task of classifying network traffic is
embarrassingly parallel once the dictionary,
embeddings, and trained classifier have been
distributed.

\begin{table}
\begin{minipage}{.5\linewidth}
\caption{Theoretical Speedup}
\label{table:speedup}
\centering
\begin{tabular}{|c| c c c|}
\hline
 & Parallel & Serial & Max\\
 & Portion & Portion & Speedup\\
\hline
 Current & 106.4 & 56.5 & 2.9 \\
 Future & 145.2 & 17.7 & 9.2 \\
\hline
\end{tabular}
\end{minipage}
\begin{minipage}{.5\linewidth}
\caption{Testing Throughput}
\label{table:throughput}
\centering
\begin{tabular}{|c| c r c r |}
\hline
 & Num & Time & Size& Rate \\
 & Files & (hours) & (GBs) & (MB/s) \\
\hline
Data creation & 600 & 15.5 & 559 & 10.3 \\
RFC 10& 600 & 14.5 & 559 & 11.0 \\
RFC 820& 300 & 25.5 & 279 & 3.1 \\
Naive Bayes & 300 & 7.6 & 279 & 10.5 \\
\hline
\end{tabular}
\end{minipage}
\end{table}

We also did some longer runs of applying 
a classifier to large sets of pcaps to gauge average
throughput.  Table \ref{table:throughput} summarizes 
the results.
For the simpler models, we can classify at about 
10.5-11 MB/s while 
packet data is created at an average rate of 10.3 MB/s.  
The original data (the first day of 
DARPA-2009) comprises 15.5 hours and 600 files.
We ran the Random Forest Classifier trained on one pcap
on the entire data set.  We also ran another Random
Forest Classifier that was trained on 300 files
and tested it on the other 300 files.  
Similarly, we trained a Naive Bayes classifier
on 300 files and tested on the other half.
For all the runs we utilized ten threads.

The difference between the two Random Forest Classifiers
is that the one trained on one pcap file has ten
estimators while the one trained on 300 files has
820 estimators.  The difference comes from the fact
that in order to incorporate knowledge from other
files to an existing Random Forest Classifier, we
had to increase the number of estimators, essentially
creating additional trees for each file.  Thus,
the Random Forest with 820 estimators has a much
lower throughput because the longer predictions times
(about 6 seconds versus 214 seconds).  In the future,
we plan to parallelize the prediction for loop which
will likely make the difference in throughput less drastic.
The Random Forest Classifier with ten estimators 
and the Naive Bayes were able to keep pace with the 
data creation rate.

\subsection{Classifier Performance}

We tested out two classifiers, the Random Forest Classifier
\cite{random-forest-2001} and Gaussian Naive Bayes
\cite{naive-bayes-1982}.  We split the first day of DARPA-2009
into two sets of 300 files, one for training and one for testing.
We listed all 600 files and gave training the even files
and testing the odd files.  This gave both sets
representative data throughout the day.  

We report two metrics, the area under the curve (AUC)
for both the Receiver Operating Characteristic (ROC) curve
and the Precision/Recall curve.  The ROC curve plots
true positive rate against the false positive rate
as the threshold is varied.  A perfect score for
the AUC is 1.0.  The ROC is known to provide overly
optimistic results when data skew is present, as is with
DARPA-2009.  

\begin{figure*}
\begin{minipage}{0.5\textwidth}
\centering
\includegraphics[width=.8\linewidth]{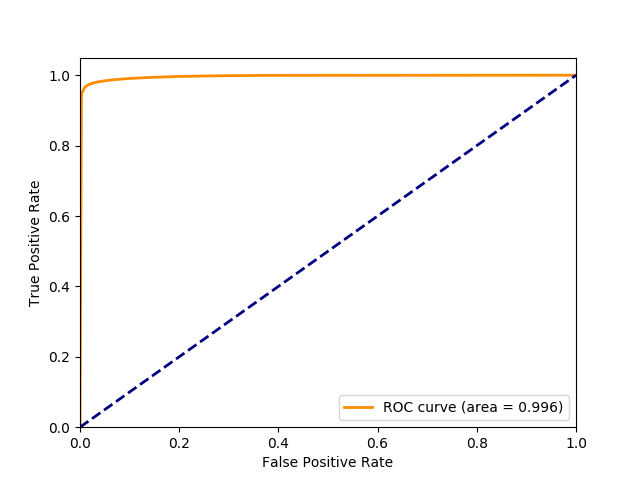}
\caption{Random Forest Classifier - Receiver Operating Characteristic}
\label{figure:roc-rfc}
\end{minipage}
\begin{minipage}{0.5\textwidth}
\centering
\includegraphics[width=.8\linewidth]{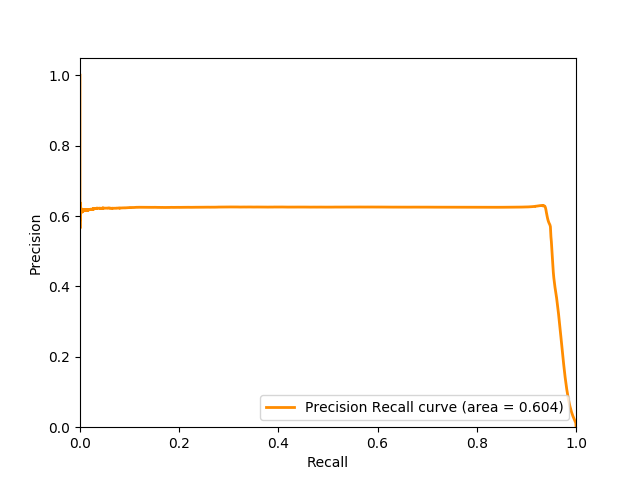}
\caption{Random Forest Classifier - Precision/Recall}
\label{figure:pr-rfc}
\end{minipage}
\begin{minipage}{0.5\textwidth}
\centering
\includegraphics[width=.8\linewidth]{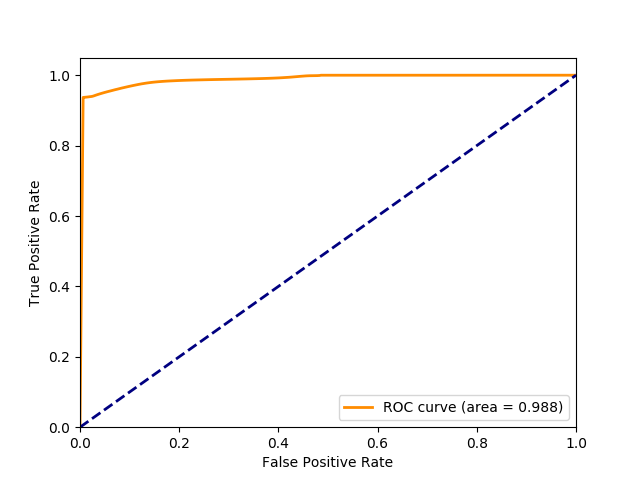}
\caption{Gaussian Naive Bayes - Receiver Operating Characteristic}
\label{figure:roc-gnb}
\end{minipage}
\begin{minipage}{0.5\textwidth}
\centering
\includegraphics[width=.8\linewidth]{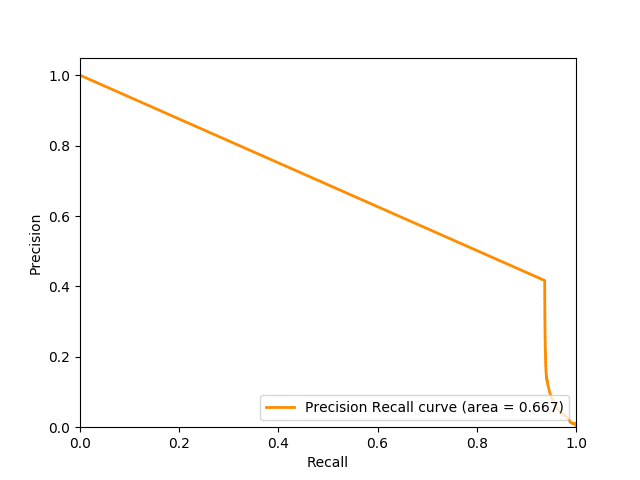}
\caption{Gaussian Naive Bayes - Precision/Recall}
\label{figure:pr-gnb}
\end{minipage}
\end{figure*}

The Precision/Recall curve emphasizes how 
good the predictions are for the minority class 
(i.e. malicious traffic).  Precision is defined
as the true positives divided by the 
true positives and false positives.  So it is 
the fraction of results that are correct returned by
the model:
$Precision = \frac{TP}{TP + FP}$.
Recall is defined as the number of true positives
divided by the true positives plus the false negatives:
$Recall = \frac{TP}{TP + FN}$.
This gives you the fraction of the entire target
class that are being returned by the model.

Table \ref{table:classifiers}
gives an overview of both classifiers and both
metrics.  The AUC ROC metric gives
a somewhat optimistic impression of the 
classifier's skill, with values between 0.988
and 0.996, while the AUC or the precision/recall
curve range between 0.604 and 0.667. 
The AUC of the precision/recall 
curve is probably more useful as it
gives an idea of how good the classifier
does at predicting the minority class.
Figures \ref{figure:roc-rfc} and \ref{figure:pr-rfc}
present the ROC and precision/recall curves
for the Random Forest Classifier, respectively,
while Figure \ref{figure:roc-gnb} and \ref{figure:pr-gnb}
are for Gaussian Naive Bayes.

\begin{table}
\caption{Classifier performance}
\label{table:classifiers}
\centering
\begin{tabular}{| c | c | c| }
\hline
 & AUC ROC & AUC Precision/Recall \\
\hline
Random Forest Classifier & 0.996 & 0.604 \\
Gaussian Naive Bayes & 0.988 & 0.667 \\
\hline
\end{tabular}
\end{table}

In both cases, there is a signficant change in the 
precision/recall curve when recall is about 0.94.  For
Gaussian Naive Bayes, the plot is a little deceptive
as the first point from the data is $(0.937, 0.417)$
with a threshold of 1, meaning that any prediction less
than one was considered benign.  The point at $(0, 1)$
is by definition.  We believe there is a large class
of malicious behavior, likely the DDoS traffic, that
both classifiers have a relatively easy time 
predicting.  The transition at $recall=0.94$ is likely
for the other classes of malicious behavior.  

Tables \ref{table:pr-rfc} and \ref{table:pr-gnb} 
provide some points along the precious/recall
curve for the two classifiers, along with the corresponding
F1 score.  This is to give an idea of the tradeoff
between finding malicious behavior and dealing
with false positives.  For instance, the first row
of Table \ref{table:pr-rfc} shows that the 
Random Forest Classifier can find 98.8\% of 
the malicious traffic, but you have to deal with 
about 94\% of the returned results being false
positives.  If that is too many, one could
use the threshold of the third line, where
about half of the returned results are actually
malicious and you still catch 95\% of the total
malicious behavior. 

\begin{table}
\begin{minipage}{.5\linewidth}
\captionsetup{width=0.95\linewidth}
\caption{Random Forest Classifier - Precision Recall}
\centering
\label{table:pr-rfc}
\begin{tabular}{| c | c | c | c|}
\hline
Precision & Recall & Threshold & F1 \\
\hline
0.060 & 0.988 & 0.006 & 0.113 \\
0.108 & 0.981 & 0.009 & 0.195 \\
0.504 & 0.951 & 0.029 & 0.659 \\
0.630 & 0.930 & 0.285 & 0.751 \\
\hline
\end{tabular}
\end{minipage}
\begin{minipage}{.5\linewidth}
\captionsetup{width=0.95\linewidth}
\caption{Gaussian Naive Bayes - Precision Recall}
\centering
\label{table:pr-gnb}
\begin{tabular}{| c | c | c | c|}
\hline
Precision & Recall & Threshold & F1 \\
\hline
0.050 & 0.963 & 2.0e-134 & 0.095 \\
0.010 & 0.947 & 1.31e-118 & 0.181 \\
0.417 & 0.937 & 0.999 & 0.577 \\
\hline
\end{tabular}
\end{minipage}
\end{table}

\section{Related Work}
\label{section:related}

A work similar to our own is that of Lotfollahi 
et al. \cite{DeepPacket2017} and their approach called
\emph{Deep Packet}.
They focus on two problems, traffic characterization (e.g. 
identifying peer-to-peer traffic) and 
application identification (e.g. identifying traffic eminating 
from Skype or Tor),
and use raw packet data as their data source.
Like our approach, they avoid hand crafted features, but
instead of a Word2Vec-based approach, they directly 
feed the packet bytes into a deep learning architecture.
Packets are truncated or padded to be 1500 bytes long, and
then fed into either a 1D convolutional neural network
or a stacked autoencoder.  


There are several papers that use deep learning, but they
apply the network to already derived features.
For the most part they test out deep learning strategies
on either KDD or NSL-KDD \cite{NSL-KDD2009, NSL-KDD2013}.
KDD is a challenge dataset from 1999 with artificially 
generated network data.  The data was composed of benign
and malicious connections, with each connection comprising
41 features.  NSL-KDD is a modification of the original KDD data
set to remove redundant records.
 
Javaid et al. \cite{Javaid:2016:DLA:2954721.2954780}
use Self-taught Learning \cite{Raina:2007:SLT:1273496.1273592}
on NSL-KDD.  Self-taught learning is an approach where you
first use an unsupervised machine learning technique
to create another representation of the data.  
For example, Javaid et al. use an autoencoder
to translate the NSL-KDD feature set into a smaller
representation.  This new representation is then 
used as the basis for classification in a supervised
training algorithm. 
Yin et al. \cite{rnn-ids-2017} also employ deep learning,
this time with recurrent neural networks, but they
also test their approach on NSL-KDD.  We agree with
the conclusions of Malowidzki et al. \cite{no-good-data-2015},
that many of the labeled public datasets are 
outdated, including NSL-KDD.

In terms of work that has examined the same data set,
Moustafa and Slay \cite{Moustafa2015} ran \emph{tcptrace}
on the first 30 files of DARPA-2009 to create flow-based
features from which they filter down to 11 features in total.
It is somewhat difficult to compare their work with ours as 
they are doing classification at the flow level, rather than
at the packet level as we do.  Also, they only examine
30 files, of which they report that 99.995\% of the malicious
activity is related to DDoS, while our 600 files covers a much
broader range of categories of malicious activity.  Also, they report
that malicious flows account for 45.5\% of their data set.
It may be a difference between flows and packets, but we 
found malicious packets to account for far less: 0.46\%.
Their best recorded model was a decision tree, that missed
10 positive examples (there were 12 total non DDoS flows)
and had no false positives.  

Ackerman et al. \cite{Ackerman2015YaleUD} also examines
DARPA-2009.  They divide up the
data into temporal chunks of one minute each, resulting
in 13,835 chunks over the ten days, with 1,848 being
malicious (if any malicious activity occured during the
time period) and 11,987 benign.  They then selected
25 features that were aggregate computations over
the time intervals.  They used diffusion maps 
\cite{diffusion-map-2005} for dimensionality reduction.
Then from a single initial point in the new feature space, 
they expand to find all similar points by recursively adding
ones that are within a certain distance of an existing point.
They do not report precision/recall numbers, but from what they
do state we calculated an average precision of 0.03 and
and average recall of 0.08, both of which are considerably lower
than our results.  However, they obtain their results from
a single example.
for finding other instances of malicious behavior in unlabeled
data.

Part of the allure of deep learning is the ability to 
extract relevant features.  Other work that focuses on
feature extraction include Ngyuen et al. \cite{Nguyen2005},
where they use sketches \cite{data-streams-book-2005}
to approximate values in the stream of network data
and Field-programmable gate arrays (FPGA's) to 
increase throughput, achieving a rate of
21.25 Gbps.  Das et al. \cite{Das2008} also use
an FPGA-based approach and a Feature Extraction Module (FEM)
based on sketches.

\section{Conclusions}
\label{section:conclusions}
We have a presented a novel application of Word2Vec, called
Packet2Vec, that translates packets into vectorized representations.
We have demonstrated promising results, with classifiers
achieving an AUC of the ROC between 0.988-0.996
and an AUC of the Precision/Recall curve between 0.604-0.667.
The method can be used on raw packet data and does 
not require any domain expertise to extract relevant features.

There are many possible avenues for future work:
\textbf{Temporal phenomenon}: We completely ignored temporal
information.  Many detection strategies utilize temporal
information to distinguish between human actors and 
bots.  How to incorporate temporal information within
a deep learning strategy for cyber data is unexplored
to our knowledge.
\textbf{Aggregating predictions}: We made classification
at the packet level.  However, to a human analyst, it 
is likely more useful to roll up predictions to the level
of a flow, or an IP, or a domain.  
\textbf{Existing features}: While we rely upon the 
deep learning model to extract relevant features, augmenting
with existing approaches could be a fecund avenue
to explore.

We believe that deep learning has much to offer cyber analysis,
and that this work is just an intial step into discovering
solutions for pressing security problems.

\section*{Acknowledgment}
Sandia National Laboratories is a multimission laboratory managed and operated by National Technology \& Engineering Solutions of Sandia, LLC, a wholly owned subsidiary of Honeywell International Inc., for the U.S. Department of Energy’s National Nuclear Security Administration under contract DE-NA0003525.

{
\bibliographystyle{splncs04}
\bibliography{bib}

\begin{thebibliography}{10}
\providecommand{\url}[1]{\texttt{#1}}
\providecommand{\urlprefix}{URL }
\providecommand{\doi}[1]{https://doi.org/#1}

\bibitem{tensorflow2015-whitepaper}
Abadi, M., Agarwal, A., Barham, P., Brevdo, E., Chen, Z., Citro, C., Corrado,
  G.S., Davis, A., Dean, J., Devin, M., Ghemawat, S., Goodfellow, I., Harp, A.,
  Irving, G., Isard, M., Jia, Y., Jozefowicz, R., Kaiser, L., Kudlur, M.,
  Levenberg, J., Man\'{e}, D., Monga, R., Moore, S., Murray, D., Olah, C.,
  Schuster, M., Shlens, J., Steiner, B., Sutskever, I., Talwar, K., Tucker, P.,
  Vanhoucke, V., Vasudevan, V., Vi\'{e}gas, F., Vinyals, O., Warden, P.,
  Wattenberg, M., Wicke, M., Yu, Y., Zheng, X.: {TensorFlow}: Large-scale
  machine learning on heterogeneous systems (2015),
  \url{https://www.tensorflow.org/}, software available from tensorflow.org

\bibitem{boost-python}
Abrahams, D., Grosse-Kunstleve, R.W.: Building hybrid systems with boost.python
  (2003)

\bibitem{Ackerman2015YaleUD}
Ackerman, D.A., Averbuch, A., Silberschatz, A., Salhov, M.: Similarity
  detection via random subsets for cyber war protection in big data using
  hadoop framework (2015)

\bibitem{SURF2008}
Bay, H., Ess, A., Tuytelaars, T., Gool, L.V.: Speeded-up robust features
  (surf). Computer Vision and Image Understanding  \textbf{110}(3),  346 -- 359
  (2008). \doi{https://doi.org/10.1016/j.cviu.2007.09.014},
  \url{http://www.sciencedirect.com/science/article/pii/S1077314207001555},
  similarity Matching in Computer Vision and Multimedia

\bibitem{random-forest-2001}
Breiman, L.: Random forests. Machine Learning  \textbf{45}(1),  5--32 (Oct
  2001). \doi{10.1023/A:1010933404324},
  \url{https://doi.org/10.1023/A:1010933404324}

\bibitem{sklearn-api}
Buitinck, L., Louppe, G., Blondel, M., Pedregosa, F., Mueller, A., Grisel, O.,
  Niculae, V., Prettenhofer, P., Gramfort, A., Grobler, J., Layton, R.,
  VanderPlas, J., Joly, A., Holt, B., Varoquaux, G.: {API} design for machine
  learning software: experiences from the scikit-learn project. In: ECML PKDD
  Workshop: Languages for Data Mining and Machine Learning. pp. 108--122 (2013)

\bibitem{naive-bayes-1982}
Chan, T.F., Golub, G.H., LeVeque, R.J.: Updating formulae and a pairwise
  algorithm for computing sample variances. In: Caussinus, H., Ettinger, P.,
  Tomassone, R. (eds.) COMPSTAT 1982 5th Symposium held at Toulouse 1982. pp.
  30--41. Physica-Verlag HD, Heidelberg (1982)

\bibitem{diffusion-map-2005}
Coifman, R.R., Lafon, S., Lee, A.B., Maggioni, M., Nadler, B., Warner, F.,
  Zucker, S.W.: Geometric diffusions as a tool for harmonic analysis and
  structure definition of data: Diffusion maps. Proceedings of the National
  Academy of Sciences  \textbf{102}(21),  7426--7431 (2005).
  \doi{10.1073/pnas.0500334102}, \url{https://www.pnas.org/content/102/21/7426}

\bibitem{Das2008}
Das, A., Nguyen, D., Zambreno, J., Memik, G., Choudhary, A.: An fpga-based
  network intrusion detection architecture. IEEE Transactions on Information
  Forensics and Security  \textbf{3}(1),  118--132 (March 2008).
  \doi{10.1109/TIFS.2007.916288}

\bibitem{DARPA2009}
Gharaibeh, M., Papadopoulos, C.: Darpa-2009 intrusion detection dataset report.
  Tech. rep., Colorado State University (2014)

\bibitem{NCE-2012}
Gutmann, M.U., Hyv\"{a}rinen, A.: Noise-contrastive estimation of unnormalized
  statistical models, with applications to natural image statistics. J. Mach.
  Learn. Res.  \textbf{13}(1),  307--361 (Feb 2012),
  \url{http://dl.acm.org/citation.cfm?id=2503308.2188396}

\bibitem{Javaid:2016:DLA:2954721.2954780}
Javaid, A., Niyaz, Q., Sun, W., Alam, M.: A deep learning approach for network
  intrusion detection system. In: Proceedings of the 9th EAI International
  Conference on Bio-inspired Information and Communications Technologies
  (Formerly BIONETICS). pp. 21--26. BICT'15, ICST (Institute for Computer
  Sciences, Social-Informatics and Telecommunications Engineering), ICST,
  Brussels, Belgium, Belgium (2016). \doi{10.4108/eai.3-12-2015.2262516},
  \url{http://dx.doi.org/10.4108/eai.3-12-2015.2262516}

\bibitem{NIPS2012_4824}
Krizhevsky, A., Sutskever, I., Hinton, G.E.: Imagenet classification with deep
  convolutional neural networks. In: Pereira, F., Burges, C.J.C., Bottou, L.,
  Weinberger, K.Q. (eds.) Advances in Neural Information Processing Systems 25,
  pp. 1097--1105. Curran Associates, Inc. (2012),
  \url{http://papers.nips.cc/paper/4824-imagenet-classification-with-deep-convolutional-neural-networks.pdf}

\bibitem{DeepPacket2017}
Lotfollahi, M., Zade, R.S.H., Siavoshani, M.J., Saberian, M.: Deep packet: {A}
  novel approach for encrypted traffic classification using deep learning. CoRR
   \textbf{abs/1709.02656} (2017), \url{http://arxiv.org/abs/1709.02656}

\bibitem{SIFT1999}
Lowe, D.G.: Object recognition from local scale-invariant features. In:
  Proceedings of the Seventh IEEE International Conference on Computer Vision.
  vol.~2, pp. 1150--1157 vol.2 (Sept 1999). \doi{10.1109/ICCV.1999.790410}

\bibitem{no-good-data-2015}
MaÅ‚owidzki, M., Berezinski, P., Mazur, M.: Network intrusion detection:
  Half a kingdom for a good dataset (04 2015)

\bibitem{word2vec-2013-1}
Mikolov, T., Chen, K., Corrado, G., Dean, J.: Efficient estimation of word
  representations in vector space. CoRR  \textbf{abs/1301.3781} (2013),
  \url{http://arxiv.org/abs/1301.3781}

\bibitem{word2vec-2013-2}
Mikolov, T., Sutskever, I., Chen, K., Corrado, G.S., Dean, J.: Distributed
  representations of words and phrases and their compositionality. In: Burges,
  C.J.C., Bottou, L., Welling, M., Ghahramani, Z., Weinberger, K.Q. (eds.)
  Advances in Neural Information Processing Systems 26, pp. 3111--3119. Curran
  Associates, Inc. (2013),
  \url{http://papers.nips.cc/paper/5021-distributed-representations-of-words-and-phrases-and-their-compositionality.pdf}

\bibitem{Moustafa2015}
Moustafa, N., Slay, J.: Creating novel features to anomaly network detection
  using darpa-2009 data set. In: 14th European Conference on Cyber Warfare and
  Security (2015)

\bibitem{data-streams-book-2005}
Muthukrishnan, S.: Data streams: Algorithms and applications. Found. Trends
  Theor. Comput. Sci.  \textbf{1}(2),  117--236 (Aug 2005).
  \doi{10.1561/0400000002}, \url{http://dx.doi.org/10.1561/0400000002}

\bibitem{Nguyen2005}
Nguyen, D., Memik, G., Memik, S.O., Choudhary, A.: Real-time feature extraction
  for high speed networks. In: International Conference on Field Programmable
  Logic and Applications, 2005. pp. 438--443 (Aug 2005).
  \doi{10.1109/FPL.2005.1515761}

\bibitem{DGX}
Nvidia dgx-1 datasheet (2017),
  \url{http://images.nvidia.com/content/technologies/deep-learning/pdf/Datasheet-DGX1.pdf},
  accessed: 2017-08-18

\bibitem{Raina:2007:SLT:1273496.1273592}
Raina, R., Battle, A., Lee, H., Packer, B., Ng, A.Y.: Self-taught learning:
  Transfer learning from unlabeled data. In: Proceedings of the 24th
  International Conference on Machine Learning. pp. 759--766. ICML '07, ACM,
  New York, NY, USA (2007). \doi{10.1145/1273496.1273592},
  \url{http://doi.acm.org/10.1145/1273496.1273592}

\bibitem{NSL-KDD2013}
Revathi, S., Malathi, A.: A detailed analysis on nsl-kdd dataset using various
  machine learning techniques for intrusion detection. International Journal of
  Engineering Research \& Technology (IJERT)  \textbf{2},  1848--1853 (01 2013)

\bibitem{istr-2018}
Symantec: Internet security threat report (2018)

\bibitem{NSL-KDD2009}
Tavallaee, M., Bagheri, E., Lu, W., Ghorbani, A.A.: A detailed analysis of the
  kdd cup 99 data set. In: 2009 IEEE Symposium on Computational Intelligence
  for Security and Defense Applications. pp.~1--6 (July 2009).
  \doi{10.1109/CISDA.2009.5356528}

\bibitem{rnn-ids-2017}
Yin, C., Zhu, Y., Fei, J., He, X.: A deep learning approach for intrusion
  detection using recurrent neural networks. IEEE Access  \textbf{5},
  21954--21961 (2017). \doi{10.1109/ACCESS.2017.2762418}

\end{thebibliography}
}

\end{document}